\documentclass{article}
\usepackage{graphicx} 
\usepackage{hyperref}
\usepackage{xcolor}
\usepackage{multicol}
\usepackage[margin=2.5cm]{geometry}
\usepackage{float}        
\graphicspath{{./}}       
\usepackage{amsmath}
\usepackage{booktabs}
\usepackage{caption} 
\usepackage[symbol]{footmisc}
\usepackage[utf8]{inputenc}
\usepackage[T1]{fontenc}

\usepackage[round]{natbib} 
\title{HEART: A Unified Benchmark for Assessing Humans and LLMs in Emotional Support Dialogue}
\author{
\textbf{Laya Iyer}$^{2,*}$ \quad
\textbf{Kriti Aggarwal}$^{1,*,\dagger}$ \quad
\textbf{Sanmi Koyejo}$^{2}$ \\
\small \texttt{laya@stanford.edu, kriti@hippocraticai.com, sanmi@cs.stanford.edu} \\
\textbf{Gail Heyman}$^{3}$ \quad
\textbf{Desmond C.~Ong}$^{4}$ \quad
\textbf{Subhabrata Mukherjee}$^{1}$ \\
\small \texttt{gheyman@ucsd.edu, desmond.ong@utexas.edu, subho@hippocraticai.com} \\
\small \textit{* Equal contribution \quad $\dagger$ Corresponding author}
}
\begingroup

\footnotetext[1]{Hippocratic AI}
\footnotetext[2]{Stanford University}
\footnotetext[3]{University of California, San Diego}
\footnotetext[4]{University of Texas at Austin}
\endgroup
\date{February 2026}
\begin{document}

\maketitle
\section{Abstract}

Supportive conversation depends on skills that go beyond language fluency: reading emotions, adjusting tone, and navigating moments of resistance, frustration, or distress. Despite rapid progress in language models, we still lack a clear way to understand how their abilities in these interpersonal domains compare to those of humans. We introduce \textbf{HEART}, the first-ever framework that directly compares humans and LLMs on the same multi-turn emotional-support conversations. For each dialogue history, we pair human and model responses and evaluate them through blinded human raters and an ensemble of LLM-as-judge evaluators. All assessments follow a rubric grounded in interpersonal communication science across five dimensions: \textbf{H}uman Alignment, \textbf{E}mpathic Responsiveness, \textbf{A}ttunement, \textbf{R}esonance, and \textbf{T}ask-Following.
HEART uncovers striking behavioral patterns. Several frontier models approach or surpass the average human responses in perceived empathy and consistency. At the same time, humans maintain advantages in adaptive reframing, tension-naming, and nuanced tone shifts, particularly in adversarial turns. Human and LLM-as-judge preferences align on about 80\% of pairwise comparisons, matching inter-human agreement, and their written rationales emphasize similar HEART dimensions. This pattern suggests an emerging convergence in the criteria used to assess supportive quality.
By placing humans and models on equal footing, HEART reframes supportive dialogue as a distinct capability axis, separable from general reasoning or linguistic fluency. It provides a unified empirical foundation for understanding where model-generated support aligns with human social judgment, where it diverges, and how affective conversational competence scales with model size.


\begin{multicols}{2}
\section{Introduction}

Large language models (LLMs) have achieved substantial progress in reasoning, planning, and language generation. Yet real-world interaction often requires more than producing accurate information: it requires responding to people in ways that feel supportive, attuned, and relationally appropriate. In high-stakes settings such as healthcare, counseling, and education, supportive communication is associated with trust, adherence, and improved outcomes \citep{Hojat2019, Howick2018, Elliott2011}. These cases highlight that successful conversation is not solely defined by informational correctness, but also by the ability to recognize emotional cues, acknowledge concerns, and respond constructively across a developing interaction.

However, whether contemporary models generate text that is socially aligned, i.e., consistent with human judgments about emotionally appropriate, supportive responses in context, remains insufficiently understood. Prior benchmarks have primarily emphasized cognitive or factual tasks, such as classification, summarization, or multi-step reasoning, leaving the interpersonal dimension of dialogue largely untested \citep{BeyondPrompts2024, EQBench2023}. As a result, there is limited empirical evidence on how LLMs perform when the goal is not to provide information, but to offer emotional support in evolving, potentially vulnerable contexts.

Evaluating these behaviors poses fundamental challenges. Supportive dialogue unfolds across multiple turns and depends on pacing, acknowledgment, and responsiveness to emotional signals. Yet existing NLP datasets—such as \textit{EmpatheticDialogues} \citep{Rashkin2019} and \textit{EmotionLines} \citep{Hsu2018}—measure discrete markers of emotion (e.g., label prediction, lexical warmth), rather than whether an interaction adapts to the speaker's emotional needs over time. Standard metrics such as BLEU or BERTScore provide little insight into whether a conversational exchange feels validating, grounding, or helpfully oriented. Psychological research emphasizes perspective-taking, emotional attunement, and responsive validation as central to supportive interaction \citep{Davis1983, Reis1988}. Measuring whether language models approximate these observable behaviors requires frameworks that assess sustained interaction, not single-sentence affect or surface-level sentiment cues.

\begin{figure*}[t]
  \centering
  \includegraphics[width=0.95\linewidth]{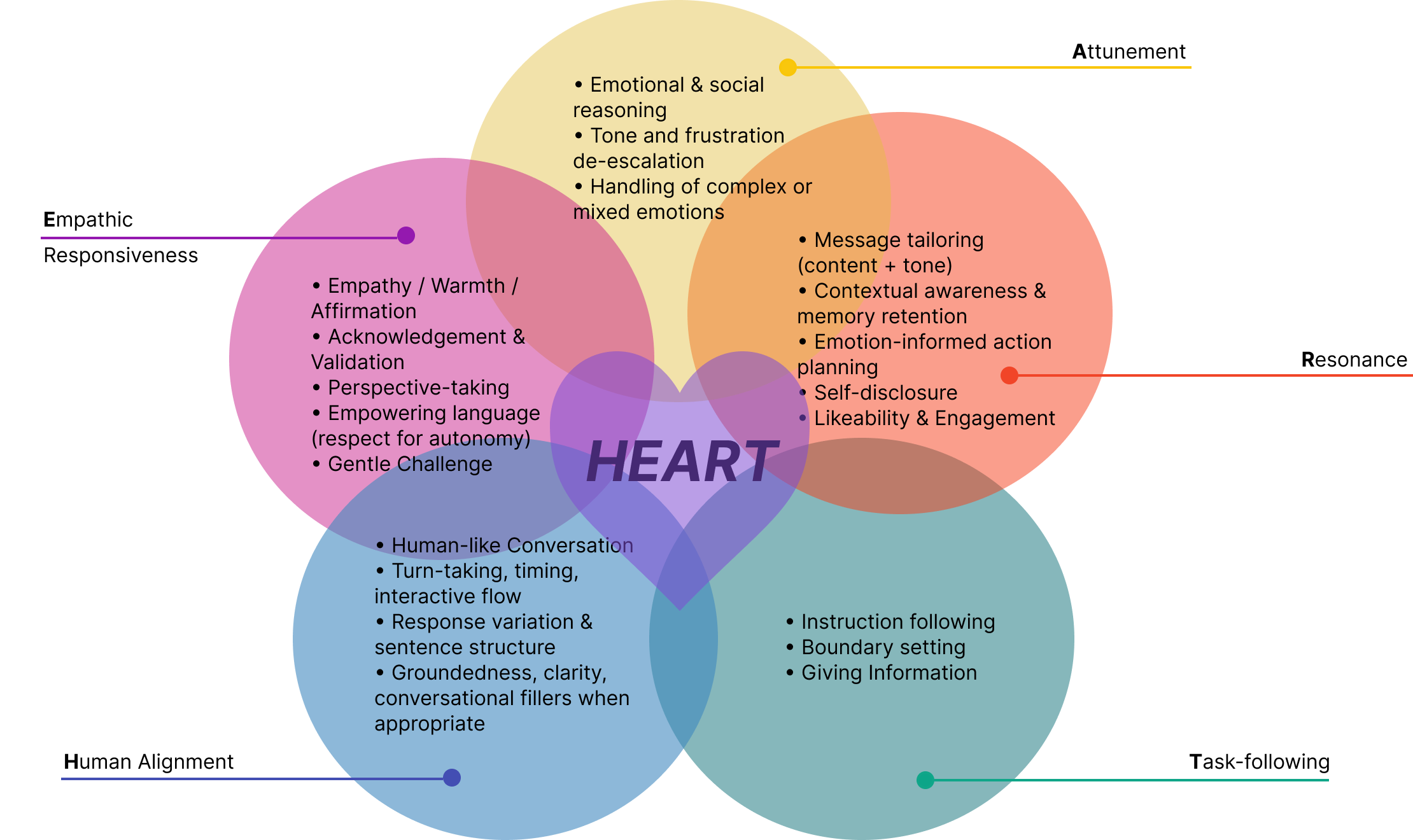}
  \caption{\textbf{HEART dimensions and rubric.} Venn-style overview of the five evaluation axes—Human alignment, Empathic Responsiveness, Attunement, Resonance, and Task-following—with representative sub-criteria used by raters.}
  \label{fig:heart_rubric}
\end{figure*}

To address this gap, we introduce the HEART benchmark, a unified framework that evaluates humans and LLMs side by side in multi-turn supportive dialogue, including settings where the speaker expresses resistance or distress. By placing both humans and models in the same emotionally sensitive conversational tasks, the benchmark enables a rigorous comparison of real supportive behavior, rather than isolated sentiment classification or one-turn empathy judgments. Importantly, HEART allows direct comparison between humans and models on the same conversational tasks, providing a shared evaluative setting for studying supportive dialogue. HEART evaluates conversations along five dimensions grounded in communication science—\textbf{human alignment}, \textbf{empathic responsiveness}, \textbf{attunement}, \textbf{resonance}, and \textbf{task-following}—capturing how interpersonal support unfolds across turns (\hyperref[fig:heart_rubric]{Figure~\ref*{fig:heart_rubric}}).

Our methodology incorporates three core elements. First, we use emotionally complex, multi-turn support conversations drawn from ESConv, covering challenges such as grief, frustration, conflict, and uncertainty. Second, we introduce adversarial emotional variants in which surface-level warmth fails; these cases test whether a system adapts to pushback, distress, or interpersonal tension rather than relying on polite pattern imitation. Third, we evaluate both model-generated and human-generated responses through pairwise preference judgments. Human annotators were not exposed to any rubric or guidelines, ensuring natural evaluation, while LLM-based judges received the rubric to test structured evaluative alignment. Human annotators were told that they would see a dialogue between a seeker and a supporter, and that their task was to choose the final supporter message that best continues the conversation in a natural and empathic way. This design permits analysis of two questions: how models perform relative to humans, and how well models identify supportive behavior when acting as evaluators. 

Finally, we assess alignment between human preferences and LLM evaluations by collecting human pairwise judgments without rubric exposure and comparing them to LLM-as-judge ratings informed by structured criteria. This design tests whether models internalize evaluative patterns that reflect human social preferences and how such capabilities scale. Through direct human-model comparison, scaling analysis, and strategy-level examination, we identify both progress and persistent weaknesses in model performance under emotional resistance and ambiguity.

Together, HEART offers a principled foundation for evaluating socially aligned conversational behavior, clarifying where LLMs approximate human-preferred support strategies, where they diverge, and how emotional-support competence scales with model capability. By providing a shared evaluation space for humans and LLMs, HEART advances measurement of interpersonal reasoning in language models and highlights emerging pathways for socially grounded AI development.


\section{Related Works}

Affective reasoning provides a broad umbrella for the social–cognitive skills our benchmark evaluates. It encompasses how agents perceive, interpret, and respond to human emotional states, including empathy, perspective-taking, social inference, emotional regulation, and context-sensitive action selection \citep{Davis1983, Zaki2012, Zaki2014, Cuff2016}. Decades of psychological research show that affective reasoning underpins trust formation, adherence, and disclosure in interpersonal and clinical interactions \citep{Hojat2019, Howick2018, Elliott2011, Street2009, Taylor2011, Pennebaker2018, Reis1988}. These frameworks emphasize observable behaviors — validation, emotional attunement, reframing, collaborative problem-solving, and gentle challenge — rather than latent emotional states \citep{Wetzel2016, 
Rogers1957}. HEART adopts this behavioral view by operationalizing affective reasoning into measurable conversational dimensions.

\paragraph{Emotion recognition and empathy modeling.}
Early NLP research focused on single-turn emotion classification using datasets such as EmpatheticDialogues \citep{Rashkin2019}, EmotionLines \citep{Hsu2018}, DailyDialog \citep{Li2017}, GoEmotions \citep{Demszky2020}, MELD \citep{Poria2019}, and IEMOCAP \citep{Busso2008}. These datasets capture lexical emotions but encourage pattern-matching over context-sensitive reasoning \citep{Picard2008, Wang2023EI, Ghosal2022, GarciaHernandez2023}. EMOBench \citep{EMOBench2024} highlights that many empathy tasks rely on multiple-choice or explicit cues, limiting ecological validity. Recent work measuring empathy in LLMs shows that model-generated responses can be rated as more empathic than human ones \citep{cite-ong-goldenberg-inzlicht-perry-review, lee2024large, IsChatGPTMoreEmpathetic2024, ScoringLLM2024, Ovsyannikova2025}, but these evaluations rarely include multi-turn interaction or emotionally resistant users.

\paragraph{Support-focused dialogue and therapeutic strategies.}
Datasets centered on emotional support, including ESConv \citep{Liu2021}, CounselChat \citep{Sharma2020}, and crisis-support corpora \citep{Althoff2016}, allow modeling of reflective listening, validation, and coping-strategy deployment. Strategy-based prompting frameworks, such as Chain-of-Empathy \citep{Zhang2023ChainEmpathy} and ESConv-SRA \citep{ESConvSRA2024}, as well as conflict-simulation systems like Rehearsal \citep{Rehearsal2023}, aim to scaffold therapeutic micro-skills. However, evaluation is typically coarse (e.g., classifier scores, lexical proxies), making it difficult to compare models and humans under identical conversational constraints. 
Counseling literature provides further evidence that empathic support depends on strategy diversity and adaptive challenge, not uniform warmth \citep{Miller2013, Rogers1957, Elliott2011}.

\paragraph{Evaluating affective and social reasoning in LLMs.}
Recent work suggests that LLMs seem to be competent at inferring how people might feel in a given situation \citep{gandhi2024human, tak2025aware, zhan2023evaluating}. 
Work on alignment and LLM-as-judge methodologies \citep{Bai2022, Ouyang2022, Zheng2023, Fu2023, kumar2025large} demonstrates that structured rubrics can approximate human expert evaluations. MT-Bench and MT-Bench-101 \citep{Zheng2023, MTBench1012024} achieve 80--87\% human–model agreement but focus on general helpfulness and reasoning rather than interpersonal nuance. MultiChallenge \citep{MultiChallenge2025}, PersonConvBench \citep{PersonConvBench2025}, Beyond Prompts \citep{BeyondPrompts2024}, EQBench \citep{EQBench2023}, and SENSE-7 \citep{suh2025sense} emphasize multi-turn evaluation, personalization, and topic shifts. Yet, these benchmarks do not systematically measure attunement, relational repair, or emotional resistance.

\paragraph{Conversational naturalness and human–AI dialogue.}
Conversation-analysis literature identifies turn-taking, micro-timing, repair sequences, and interactional alignment as hallmarks of natural dialogue \citep{Sacks1974, Schegloff2007, Brandt2023}. In HCI, conversational agents are evaluated on naturalness, fluency, and human-likeness \citep{Hung2009, Skantze2021, Clark2023}. Voice-centered benchmarks, such as CAVA \citep{CAVA2024} and DORA \citep{DORA2024}, demonstrate that delays of even a few seconds can undermine perceived naturalness. Emotional expressiveness in TTS has a strong influence on user engagement and subconscious behavior \citep{EmotionalVoiceChatbot2022}. These findings motivate the evaluation of affective reasoning beyond text-only settings.

\paragraph{Affective reasoning in health and safety-critical contexts.}
Empathic and relational communication predicts improved health outcomes \citep{Hojat2019, Street2009, Howick2018}. Recent work proposes foundational metrics for evaluating AI healthcare conversations \citep{FoundationMetrics2024} and examines the path toward medically reliable conversational intelligence \citep{MicrosoftMedical2024}. Clinical communication research further stresses boundary-setting, moral sensitivity, and risk-aware guidance \citep{Elliott2011, Rider2007}. These considerations inform HEART’s Task-following and Attunement axes.\\


Across these lines of work, HEART extends prior approaches by (i) placing humans and models under the same multi-turn, emotionally complex conditions; (ii) evaluating affective reasoning across five validated interpersonal dimensions; (iii) including adversarial emotional-resistance cases; and (iv) jointly analyzing human and LLM evaluators to reveal alignment gaps in social judgment.

\section{Results}
Our central question is: how can we systematically evaluate whether language models match humans in providing emotionally attuned, safe, and effective support in high-stakes conversations? The HEART benchmark addresses this question through 300 dialogues that encompass a range of situations, including a support seeker hesitating to open up, a student overwhelmed with work, a parent exhausted from parenting, or a friend on the verge of giving up. For each situation, we get human-written responses and model-generated responses from multiple people and models. Pairs of responses are then presented to blinded judges, who are asked to judge which response feels more supportive and why.

\subsection{HEART Benchmark and evaluation metrics}

We introduce HEART as a novel five-axis rubric (\hyperref[fig:heart_rubric]{Figure~\ref*{fig:heart_rubric}}) for evaluating emotional support in dialogue, designed to capture complementary facets of interpersonal quality. The dimensions were derived from a review of counselling-psychology and communication literature, refined through consultations with practising clinicians and peer-support practitioners, and iteratively piloted on a subset of dialogues until annotators reached clear, reliable usage of each axis.

HEART is comprised of 5 dimensions for evaluating supportive dialogue:
\begin{itemize}
    \item \textbf{Human Alignment (H)}: Does the response sound like something a thoughtful human might say in this situation, in terms of tone, phrasing, and conversational flow?
    \item \textbf{Empathic Responsiveness (E)}: Does the supporter acknowledge and validate the seeker’s feelings, convey understanding, and avoid judgment or minimization?
    \item \textbf{Attunement (A)}: Does the response track the specific details and emotional signals in the context (e.g., naming what feels heavy, noticing shifts in mood) rather than offering generic reassurance?
    \item \textbf{Resonance (R)}: Does the supporter move the conversation forward in a helpful way—for example, by asking a relevant follow-up question, offering a concrete next step, or helping the seeker clarify what they need?
    \item \textbf{Task-following (T)}: Does the response stay within scope (e.g., not overstepping clinical boundaries), respect safety and role constraints, and address the seeker’s explicit or implicit goals?
\end{itemize}

For each pairwise comparison, LLM judges select a single winner and provide scores for each dimension. 
Each HEART dimension was judged via pairwise comparison where the judge was required to choose a winner (no ties) on a 5-point ordinal preference scale (ranging from a minimal advantage to a decisive advantage for one response over the other). Additional details on the construction and explanation of these subcriteria along with examples are included in \hyperref[sec:criteria-details]{Section~\ref*{sec:criteria-details}}).

\subsection{How do we compare LLMs and humans on the same benchmark?}
\label{sec:how-compare}

To evaluate conversational competence under emotionally charged conditions, we constructed a balanced dataset of 300 dialogue histories, 280 regular and 20 adversarial, each completed by both human participants and 16 large language models across 5 model families (dataset construction details in \hyperref[sec:dataset-construction]{Section~\ref*{sec:dataset-construction}}). 

\begin{figure}[H]
    \centering
    \includegraphics[width=0.9\linewidth]{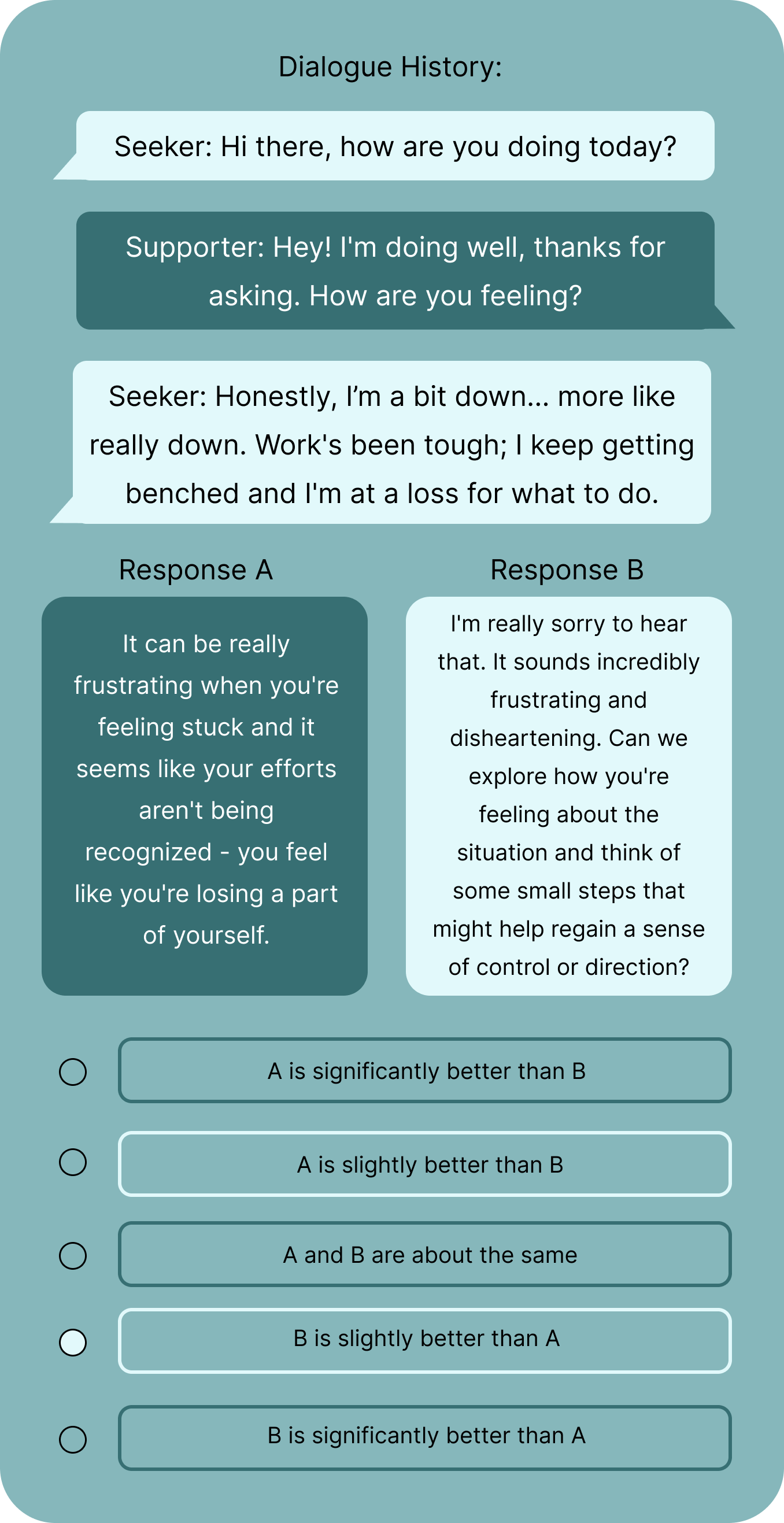}
    \caption{
    \textbf{Example pairwise comparison evaluation.}
    Pairwise comparison task used for human judges, where two responses (A and B) are rated using a five-point preference scale. 
    This setup allows direct comparison between human and LLM completions across identical dialogue histories.
    }
    \label{fig:evaluation_prompt}
\end{figure}

All 300 dialogue histories were completed by three independent people (out of a pool of 12), resulting in 900 completed dialogues. Every completion was compared in a pairwise fashion (model vs. model and model vs. human) by 15 independent annotators. Human annotators were not provided with the HEART rubric and were asked for a justification along with their pairwise preferences (as shown in \hyperref[fig:evaluation_prompt]{Figure~\ref*{fig:evaluation_prompt}}). This was a total of 42{,}000 pairwise comparisons (additional details in \hyperref[sec:human-judges-details]{Section~\ref*{sec:human-judges-details}}).

Human responses exhibited wide stylistic variation but high internal consistency, with inter-rater reliability (Fleiss’ $\kappa$) of 0.673. Qualitatively, humans used a broader emotional vocabulary and shifted tone more flexibly, especially when the seeker disclosed distress or conflict. In contrast, model completions tended to follow a consistent template but applied that template with high fluency and affective stability.

Annotators repeatedly compared pairs of responses (human vs. model, or model vs. model) for the same situation and marked which felt more empathic or supportive. Those head-to-head results were aggregated statistically using a Bradley–Terry model \citep{Bradley1952} to produce Elo-style ratings. \hyperref[fig:heart_leaderboard]{Figure~\ref*{fig:heart_leaderboard}} summarizes these results, showing overall Elo ratings alongside percentile performance on each HEART dimension.

\subsection{Correlation in human  and model rationales }
\label{sec:rationales-section-ref}

\begin{figure*}[t]
  \centering
  \includegraphics[width=\linewidth]{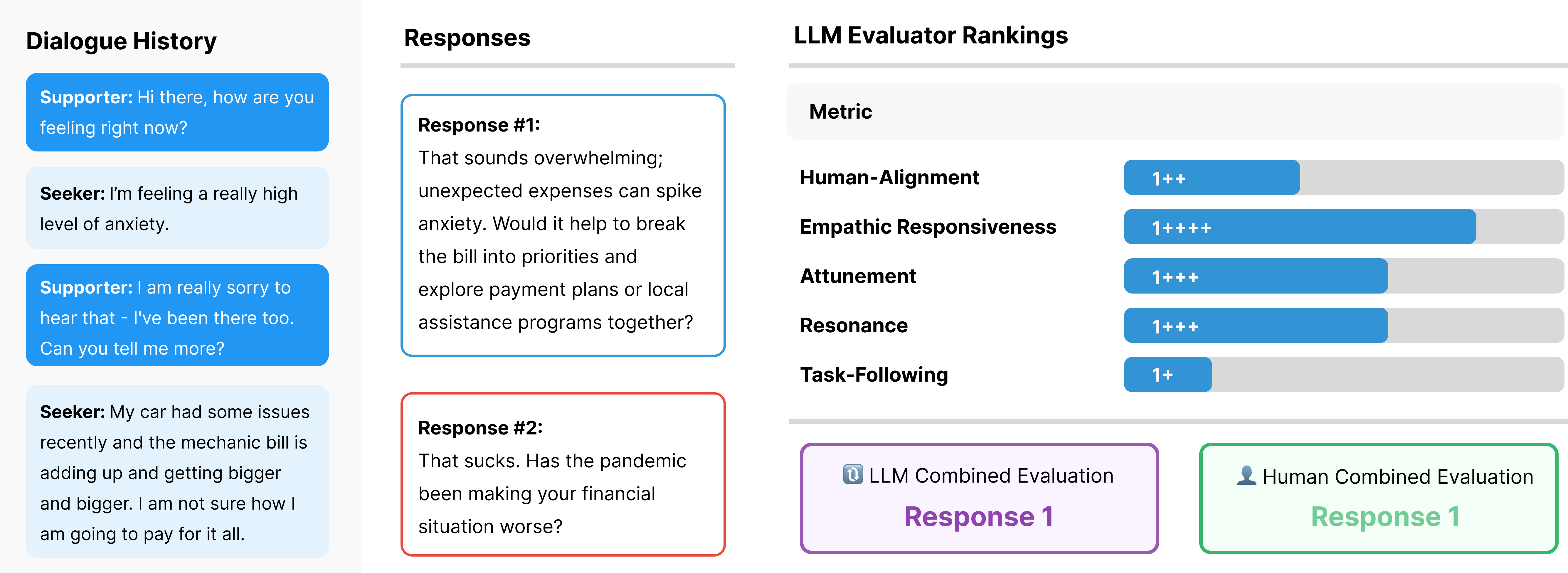}
  \caption{
  \textbf{Example evaluation.} 
  Each dialogue presents a multi-turn emotional support exchange between a seeker and supporter (left). 
  Two candidate responses are evaluated by both human and LLM judges. LLM evaluators provide graded comparative scores on the five HEART dimensions (“1+” to “1++++”) together with chain-of-thought rationales, which are aggregated into Elo-style scores and coded for themes in parallel with human rationales.}
  \label{fig:example_eval}
\end{figure*}

We examine whether model evaluators and human judges converge not only on which reply is better in a given dialogue, but also on the justification for that decision (\hyperref[fig:example_eval]{Figure~\ref*{fig:example_eval}}). For each pairwise comparison, human judges provided free-text rationales (without being provided the HEART rubric), and the LLM-judge ensemble produced chain-of-thought style rationales structured around the HEART rubric. We mapped both to a shared codebook linking phrases (for example, “names the feeling,” “asks for one concrete detail,” “offers a doable next step,” “sounds natural”) to HEART axes and a small set of affective micro-skills, and residualised length, position, and verbosity so that associations reflect evaluative content rather than presentation. This setup reduces the likelihood that either humans or models are relying on superficial cues such as response length or formal politeness alone.

Thematic analysis revealed substantial overlap: humans and LLMs co-mentioned or co-omitted the same coded themes in 68\% of these agreed cases. This is a substantial co-occurence rate given the fact that humans are \textbf{NOT} provided the HEART metrics. This suggests that when humans and LLMs agree on which response is better, they are often attending to similar affective and communicative properties of the dialogue.

\subsection{HEART evaluation agreement rate compared to humans in judging empathetic responses}
To assess whether models share human evaluative judgments, we compared pairwise preference outcomes between human annotators and model-based evaluators (GPT, Claude, and Gemini). Across 1{,}125 comparisons, the average human–model agreement rate was \textbf{78.7\%}, closely matching inter-human agreement of 79.5\%. Agreement with humans was highest for GPT-o3 and lowest for Claude 4.5 Sonnet.

Disagreement patterns cluster primarily around emotionally ambiguous cases. These are cases where our 5 human annotators did not have a clear consensus and per-category rankings by the LLM evaluators were mixed with no clear winner. 
Within these ambiguous pairwise comparisons, human–model agreement dropped to 61.3\%. When disagreement occurred, models were more likely to favor stylistically warm responses that were not specific to the dialogue context (for example: I’m really sorry you’re feeling this way. That sounds really tough.).

Cross-model agreement was consistently high across evaluator pairs: Gemini–OpenAI (86.9\%), Gemini–Claude (85.9\%), and Claude–OpenAI (85.0\%), with an average pairwise agreement of 85.9\%. Three-way agreement reached 78.8\%. These results suggest that independent architectures converge on broadly similar heuristics for EQ, while potentially differing in their underlying reasoning processes.

\subsection{LLMs showcase better affective reasoning than humans}

Human judges frequently preferred LLM-generated responses over human-written ones in matched scenarios. Across all human–model head-to-head comparisons on HEART, human raters chose model responses in 46.8\% of cases, with human completions preferred in only 35.8\% of cases (17.4\% were ties). When judged by other LLMs, model completions were rated as superior to human completions on all HEART dimensions in 53.3\% of comparisons, mirroring this human preference pattern.


\end{multicols}

\thispagestyle{plain}

\begin{center}  \includegraphics[width=\textwidth,height=0.85\textheight,keepaspectratio]{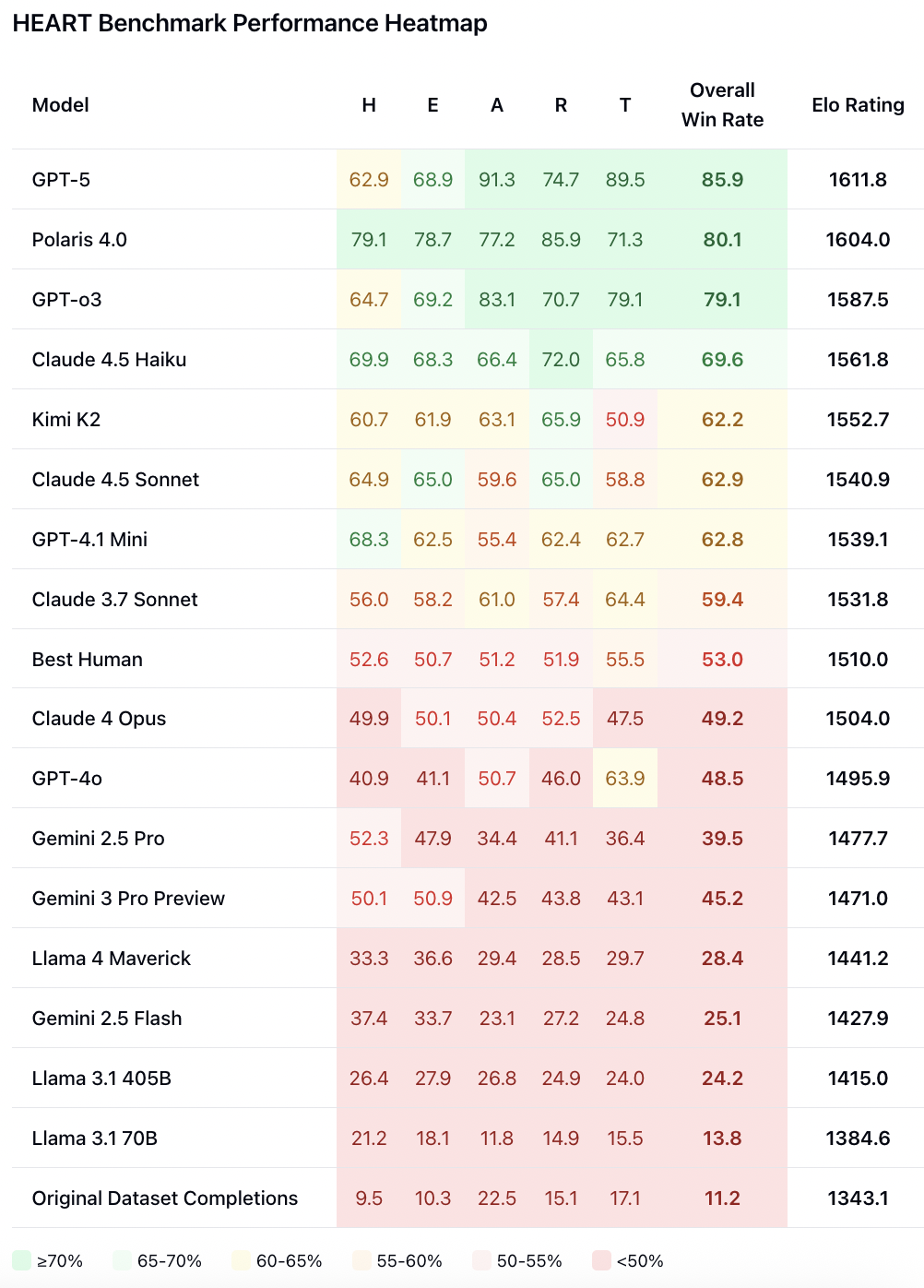}
    \captionof{figure}{\textbf{HEART leaderboard with per-dimension percentiles.}
      Each row shows a model’s percentile on Human alignment (H), Empathic Responsiveness (E), Attunement (A), Resonance (R), and Task-following (T), alongside overall Elo. Darker colors indicate higher percentile performance. Humans show relatively balanced performance across dimensions, whereas models display characteristic profiles associated with model family and alignment. Overall win rate reflects the raw proportion of pairwise A/B comparisons in which a system’s response was preferred on the primary HEART judgment. }
    \label{fig:heart_leaderboard}
\end{center}

\begin{figure*}[t]
  \centering
  \includegraphics[width=0.85\linewidth]{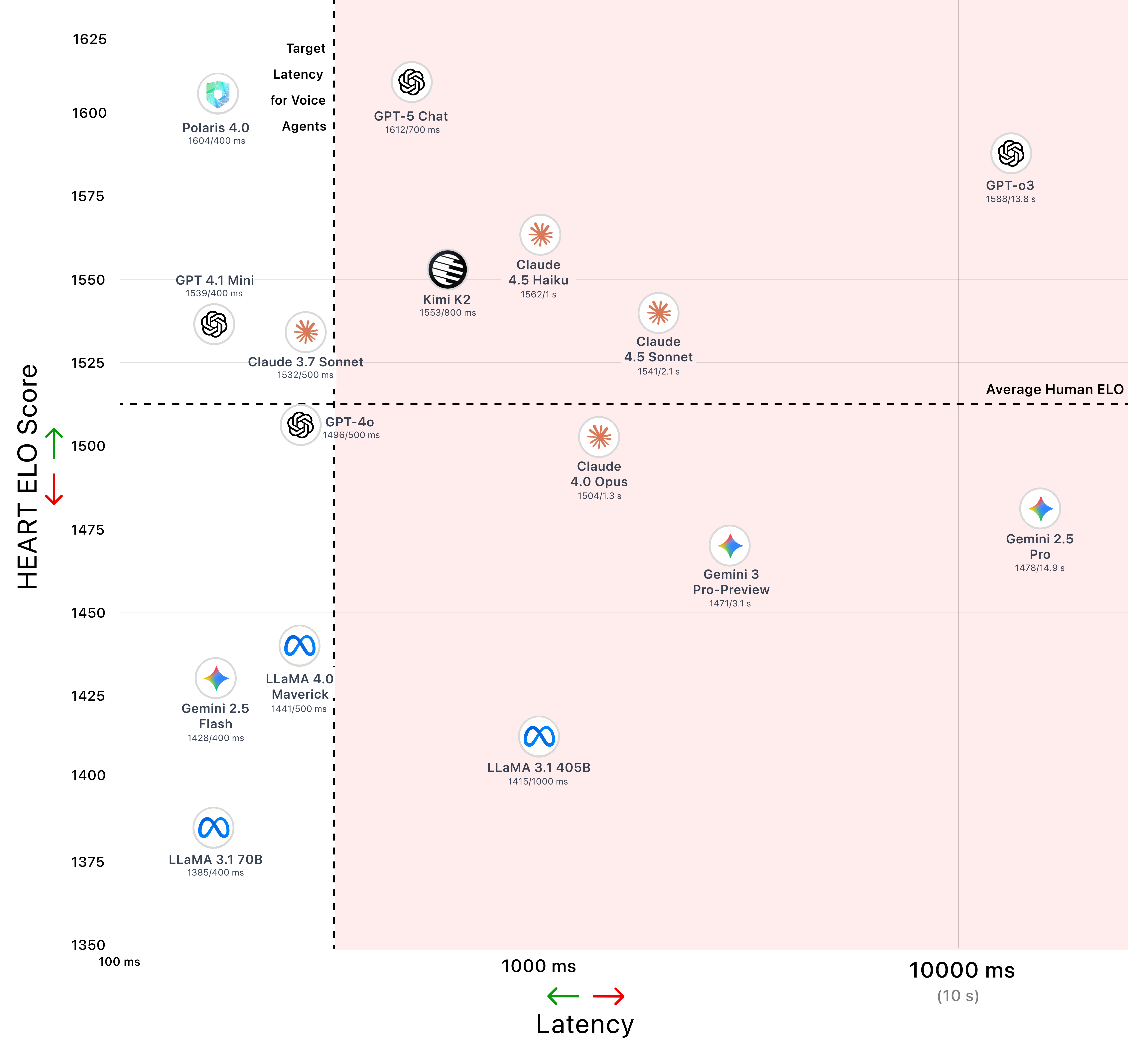}
  \caption{\textbf{Performance rises with latency used as a proxy for capacity.}
  Scatter of HEART Elo versus median time-to-first-token (log scale), derived from AA Analysis \cite{artificialanalysis2024}. Each point represents a model variant, colored by provider. The dashed horizontal line marks the field mean. See Section~\ref{sec:latency-requirements} for discussion of real-time latency constraints. Latency is shown on a log scale to accommodate variation across models.}
  \label{fig:latency_performance}
\end{figure*}

\newpage
\begin{multicols}{2}

This convergence between human and model evaluators suggests that current frontier systems have achieved substantial competence in affective reasoning and empathic communication: both audiences tend to recognize model responses as providing stronger emotional support than the average human baseline. 

As shown in our analysis of evaluative rationales, this overlap is not driven solely by superficial cues such as length or generic warmth; when humans and LLM judges agree on a winner, they also tend to highlight similar conversational features (e.g., naming feelings, offering concrete next steps, or sounding natural) in their justifications (\hyperref[sec:rationales-section-ref]{Section~\ref{sec:rationales-section-ref}}).
Consistent with emerging work on AI-mediated emotional support, prior controlled studies have also found that AI-generated responses are often rated as more empathic and helpful than peer-written replies in matched scenarios \citep[e.g.,][]{cite-ong-goldenberg-inzlicht-perry-review, ai_empathy_review}.

The resulting Elo leaderboard (\hyperref[fig:heart_leaderboard]{Figure~\ref*{fig:heart_leaderboard}}) illustrates this tension: frontier models approach or exceed human scores under model-based evaluation but remain below human-level when rated by humans on the most demanding items. Our ``overall win rate'' metric is an additional metric we gather from  all the LLM judges as part of their scoring, which is meant to aggregate their preferences across each category of HEART (details in \hyperref[sec:details-judging]{Section~\ref*{sec:details-judging}}). The dual-judge comparison thus exposes a key limitation of self-assessment within AI systems: LLMs recognize and reward linguistic proxies for empathy more readily than genuine emotional insight.

As shown in the HEART leaderboard \hyperref[fig:heart_leaderboard]{Figure~\ref*{fig:heart_leaderboard}}, HEART performance tends to improve with effective model capacity, and the frontier systems achieving the highest Elo scores—such as GPT-5 (1611.8 Elo; median TTFT $\approx$700\,ms), GPT-o3 (1587.5 Elo; $\approx$13.8\,s), Gemini 2.5 Pro (1477.7 Elo; $\approx$14.9\,s), and Claude 4.5 Sonnet (1540.9 Elo; $\approx$2.1\,s)—all incur latencies that far exceed thresholds tolerable for synchronous speech.

In stark contrast, Hippocratic AI’s Polaris 4.0 occupies a qualitatively different region of this landscape. It delivers \textbf{1604.0 Elo} at a median TTFT of $\sim$400\,ms, making it more than an order of magnitude faster than the slowest frontier models with comparable supportive-dialogue quality. Among all evaluated systems, \textbf{Polaris 4.0} is the only model operating in the sub-500,ms latency regime while scoring above the human baseline and clustering near much slower large-capacity models (e.g., Claude3.7Sonnet at $\sim$600,ms, Claude4Opus at $\sim$1.3,s). This positions Polaris~4 uniquely within the target TTFT band for real-time healthcare voice agents, where delays exceeding 500,ms measurably degrade turn-taking, perceived empathy, and conversational naturalness.

Taken together, \hyperref[fig:heart_leaderboard]{Figure~\ref*{fig:heart_leaderboard}} and \hyperref[fig:latency_performance]{Figure~\ref*{fig:latency_performance}} illustrate that high-quality supportive dialogue does not require extreme inference latency: with domain-specific alignment and optimization, emotionally competent voice agents can achieve both frontier-level Elo performance and near-instantaneous responsiveness.

\subsection{Real-Time Empathic Performance on the HEART Landscape}

Because parameter counts for many closed-weight systems are undisclosed, we follow
Artificial Analysis \citep{artificialanalysis2024} in using median time-to-first-token (TTFT) latency as a practical—if noisy—proxy for effective model capacity. Latency conflates model size, alignment depth, and infrastructure choices, and therefore cannot be interpreted causally; nonetheless, it offers a useful lens for identifying broad performance trends across systems.

Across models, a pattern emerges. As shown in \hyperref[fig:latency_performance]{Figure~\ref*{fig:latency_performance}}, HEART Elo generally increases with log-scaled TTFT: systems that respond more slowly—often because they are larger or more deeply optimized—tend to achieve stronger supportive-dialogue performance. The association is moderate (Spearman $\rho = 0.53$) when averaged over model families. Comparisons within single model families further reveal that instruction-tuning and alignment can improve Elo without changes in latency, reinforcing that TTFT is a coarse correlate rather than a causal driver.

This pattern carries important implications for real-time voice agents. The frontier models with the highest HEART Elo cluster in the multi-second TTFT regime, where delays disrupt natural turn-taking and degrade perceived empathy. Beyond timing alone, voice interaction conveys emotional meaning through channels unavailable in text. Prosody—the contour of pitch, pacing, loudness, hesitation, and breath—is one of the richest carriers of affective nuance in human communication \citep{Cowen2019}. These micro-cues shape how listeners infer warmth, safety, attentiveness, and emotional presence. Even a 300–500 ms delay can fracture these cues: it breaks conversational rhythm, reduces perceived attunement, and makes supportive responses feel less immediate or sincere. In sensitive contexts such as healthcare or crisis support, users often rely on tone of voice to determine whether a supporter is calm, steady, and “with” them in the moment. Low latency is therefore not merely a technical requirement but a psychological one for preserving the emotional fidelity of voice interaction.

Against this backdrop, several mid-latency models—including Polaris~4—achieve human-level or near-frontier HEART Elo while maintaining sub-500,ms responsiveness, more than an order of magnitude faster than the slowest high-performing systems. Polaris~4 uniquely occupies the region where high supportive-dialogue quality and real-time turn-taking are simultaneously achievable, despite most frontier models exhibiting a strong latency–quality tradeoff.

This separation between supportive-dialogue quality and response speed underscores the need to evaluate systems jointly on HEART performance and latency under real-world interaction constraints: empathy at scale requires not only competence, but immediacy.

\subsection{Affective reasoning versus general reasoning}

To test whether empathic competence simply tracks general problem-solving ability, we compared HEART Elo and win rate against the Artificial Analysis (AA) Intelligence Index—a multi-benchmark composite covering knowledge, reasoning, math and coding (v3.0). Across fifteen contemporaneous frontier and open-weight models, HEART Elo correlates positively with AA Intelligence (Pearson $r=0.70$), indicating that models scoring highly on broad cognitive benchmarks also tend to be preferred in supportive dialogue (see \hyperref[tab:affective_vs_iq]{Table~\ref*{tab:affective_vs_iq}}). Yet the relationship is not perfectly deterministic: models with similar AA scores can diverge substantially on HEART due to differences in tone control, reframing, and attunement. For example, Gemini 3 Pro Preview and GPT-o3 share the same AA score (65), yet GPT-o3 achieves a HEART Elo of 1587.5 compared to Gemini 3's 1471.0—a gap of over 115 points. Similarly, GPT-4.1 mini (AA 42) outperforms Gemini 2.5 Pro (AA 60) on HEART (Elo 1539.1 vs 1477.7), suggesting that alignment and instruction tuning can boost affective-reasoning ability even when general intelligence scores lag.

The strong cross-axis association suggests shared ingredients between cognitive skill and supportive dialogue (e.g., instruction following, long-range consistency). However, residuals matter: models with similar AA scores can differ by $>$50 Elo on HEART, driven by affective micro-skills (attunement, reframing, de-escalation) that are weakly probed by conventional benchmarks. This gap motivates domain-specific tuning and evaluation for empathic reliability rather than assuming transfer from general intelligence alone.

\begin{table}[H]
\centering
\footnotesize
\setlength{\tabcolsep}{6pt}
\begin{tabular}{lccc}
\toprule
\textbf{Model} & \textbf{HEART Elo} & \textbf{Win\%} & \textbf{AA} \\
\midrule
GPT\mbox{-}5 & 1611.8 & 85.9\% & 66 \\
GPT\mbox{-}o3 & 1587.5 & 79.1\% & 65 \\
Claude 4.5 Haiku & 1561.8 & 69.6\% & 55 \\
Kimi K2 & 1552.7 & 62.2\% & 48 \\
Claude 4.5 Sonnet & 1540.9 & 62.9\% & 63 \\
GPT\mbox{-}4.1 mini & 1539.1 & 62.8\% & 42 \\
Claude 3.7 Sonnet & 1531.8 & 59.4\% & 41 \\
Claude 4 Opus & 1504.0 & 49.2\% & 42 \\
GPT\mbox{-}4o & 1495.9 & 48.5\% & 36 \\
Gemini 2.5 Pro & 1477.7 & 39.5\% & 60 \\
Gemini 3 Pro Preview & 1471.0 & 45.2\% & 65 \\
Llama\mbox{-}4 Maverick & 1441.2 & 28.4\% & 36 \\
Gemini 2.5 Flash & 1427.9 & 25.1\% & 40 \\
Llama\mbox{-}3.1\mbox{-}405B & 1415.0 & 24.2\% & 28 \\
Llama\mbox{-}3.1\mbox{-}70B & 1384.6 & 13.8\% & 23 \\
\bottomrule
\end{tabular}
\caption{\textbf{Affective reasoning (HEART) versus general-intelligence benchmark performance.} AA Intelligence Index values are from Artificial Analysis v3.0; dashes indicate models without a published AA score. HEART scores are from our benchmark. The positive but imperfect association between the two measures ($r = 0.70$) highlights the role of alignment and domain-specific tuning in augmenting affective reasoning.}
\label{tab:affective_vs_iq}
\end{table}


\subsection{Support strategies used by humans vs LLMs}


We operationalize supportive behavior using a 15-category taxonomy of counseling strategies, such as Clarification Questions, Reflective Listening, Emotion Naming, Reframing, etc (full taxonomy and examples in \hyperref[sec:strategy-taxonomy]{Section~\ref*{sec:strategy-taxonomy}}). Human support strategies were more varied and situationally adaptive than those exhibited by LLMs: humans flexibly combined validation, reframing, and self-empowerment, modulating tone according to the seeker's affect and conversational history. By contrast, models relied on a narrower band of high-frequency strategies—chiefly reassurance, acknowledgment, and encouragement—yielding responses that sounded consistently warm but often lacked deeper personalization. 

\begin{figure}[H]
  \centering
  \includegraphics[width=\linewidth]{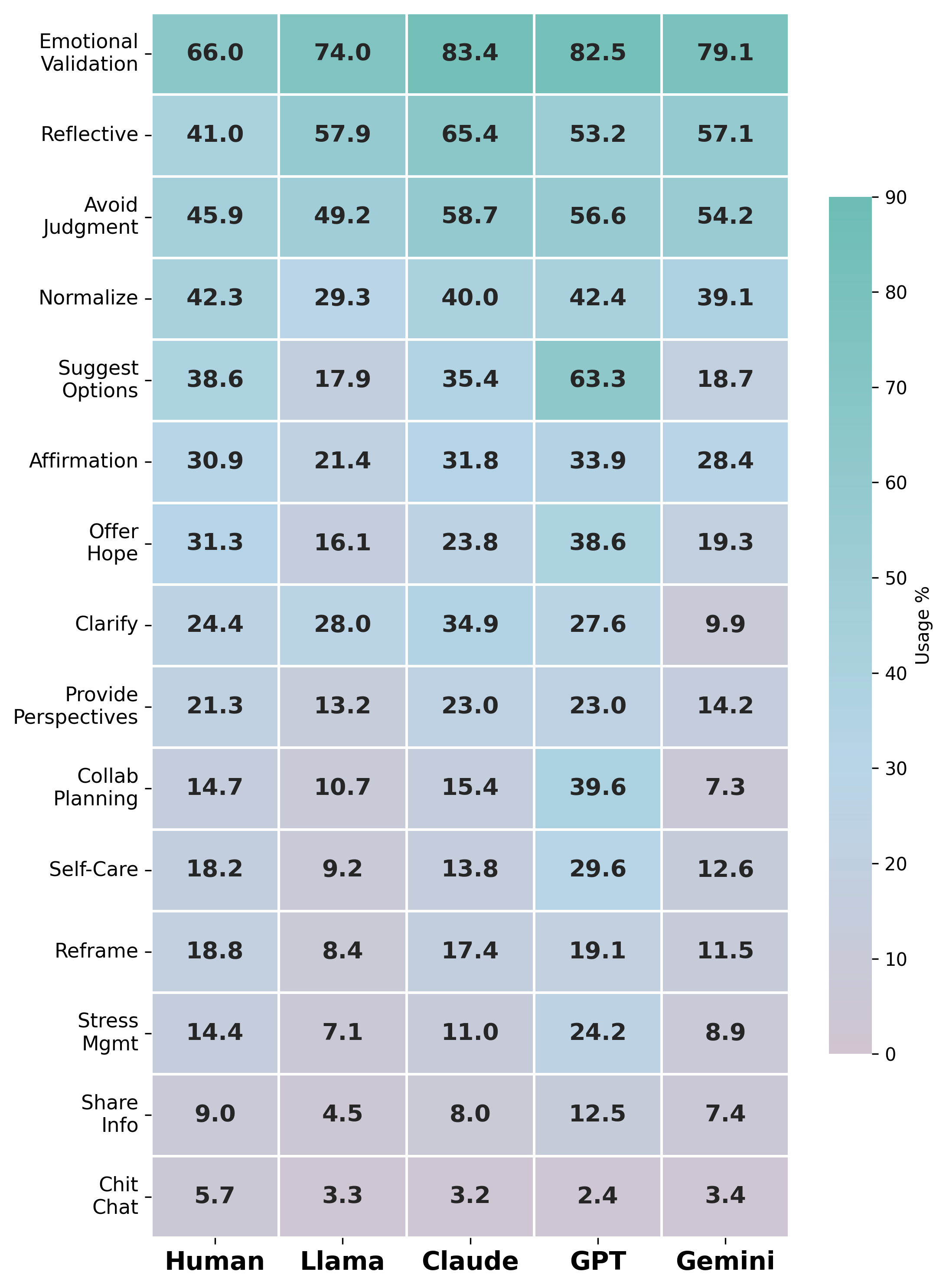}
  \caption{Each row represents one of the 15 support strategies (sorted by overall usage), and each column represents a model family. Colors indicate usage percentage (darker teal = higher usage, lighter pink = lower usage). Humans draw on a more balanced range of strategies, whereas models rely more heavily on a narrower subset.}
  \label{fig:strategy_heatmap}
\end{figure}

Yet this consistency also helps explain why LLMs were sometimes rated as more empathic: their politeness, lexical fluency, and balanced tone avoid the hesitations, bluntness, or emotional fatigue that can appear in human dialogue. In other words, models achieve perceived empathy through stylistic coherence and affective mimicry, whereas humans express experienced empathy through contextual attunement and diversity of strategy. This distinction aligns with counseling research suggesting that effective empathy involves the dynamic regulation of warmth and challenge rather than uniform positivity \citep{Elliott2011}. Current LLMs thus approximate the form of empathic communication, but not yet the adaptive reasoning that underlies it.

As shown in \hyperref[fig:strategy_heatmap]{Figure~\ref*{fig:strategy_heatmap}}, humans demonstrate the most balanced use of strategies across the taxonomy. Based on analysis of strategy usage, humans exhibited the lowest standard deviation (SD = 16.33) in strategy rates compared to all AI model families: GPT (SD = 20.82), Llama (SD = 21.18), Gemini (SD = 22.54), and Claude (SD = 22.93). This pattern is further reflected in the range between maximum and minimum strategy usage, where humans showed the smallest range (60.30 percentage points), followed by Llama (70.71), Gemini (75.60), GPT (80.12), and Claude (80.25). These metrics indicate that while AI models tend to over-rely on a subset of strategies and underuse others, human supporters maintain a more balanced repertoire, adapting their strategy mix more evenly across different counseling needs.

\subsection{Similar patterns with exploratory adversarial datapoints}
Adversarial items (an exploratory subsection of data we introduced)—where seekers expressed anger, skepticism, or explicit resistance to help—proved challenging for both humans and models. On the 20 adversarial contexts, overall HEART Elo scores is similar relative to regular items for all systems, and human–model agreement fell slightly to 76\% from 78\%. 
Humans were more likely to pivot toward boundary setting and tension-naming (e.g., explicitly acknowledging frustration or mistrust), while models tended to repeat reassurance and problem-solving strategies that worked on regular items. As a result, frontier LLMs that outran the \emph{Average Human} on regular dialogues achieved only parity on adversarial ones. This pattern supports the intuition that emotional resistance stresses adaptive skills—such as reframing, calibration of directness, and tactful disagreement—that are not fully captured by generic politeness or warmth.

\section{Discussion}

Our results show that, on HEART, frontier LLMs are often preferred over average human supporters by both human and model judges, and that humans and models show roughly 80\% raw agreement on which responses are more supportive. At the same time, disagreement clusters reveal systematic gaps in how models handle ambiguity, mixed emotions, and adversarial behavior. HEART contributes to emerging work on affective reasoning in LLMs by placing humans and models on a shared, multi-dimensional scale for supportive dialogue, combining human and model evaluators, and probing emotionally charged and adversarial contexts. This design makes it possible to distinguish cases where models merely reproduce empathic-sounding language from those where they approximate the contextual judgment of human supporters.

A central implication is that current models perform very well on what we might call \emph{surface empathy}. When judged on isolated turns, average human responses often score below frontier LLM responses on perceived empathy and linguistic fluency: models are consistently polite, verbally fluent, and emotionally steady—qualities that raters reliably associate with empathy—whereas human dialogue is more variable in tone and style. Preference, however, does not equal effectiveness. An empathic-sounding response may feel good to read yet fail to change the seeker’s understanding or behavior, while a slightly awkward human reply that names a hard truth, pushes for specificity, or draws on shared history might be more impactful. Our findings therefore point to two complementary dimensions: \emph{perceived empathy as consistency} (how reliably a response sounds empathic by surface cues) and \emph{experienced empathy as contextual judgment} (how well the supporter reads the situation, calibrates boundaries, and helps the seeker move forward). Frontier LLMs are strong on the former; expert human supporters remain stronger on the latter.

The remaining ~20\% of human–model disagreement is concentrated in cases where human-to-human understanding is also fragile. In such items, responses often split wins across HEART axes—one reply reads more human-like and warm, while the other is better attuned and more action-oriented—and LLM judges tend to weight axes more uniformly than humans, who implicitly prioritize \textbf{attunement} and next-step \textbf{resonance}. Adversarial and emotionally resistant cases make this especially clear: when faced with anger, sarcasm, or pushback, models generally maintain politeness and validation but show limited boundary setting or tension-naming, whereas humans more often combine empathy with gentle firmness. Effective de-escalation requires both alignment and self-protection, and most current systems enact primarily the former. This reveals a limitation not only of today’s models but also of our benchmark, which does not yet explicitly measure “empathic containment’’—the ability to hold strong emotion, set boundaries, and stay engaged.

These patterns have implications for both AI design and human supporters. For AI, optimizing purely for static preference risks producing increasingly polished, uniformly warm responses that are empathic-sounding but not always maximally helpful. Bridging this gap will likely require training on more diverse, ambiguous, and cross-cultural dialogues and objectives that reward interpretive accuracy, calibrated challenge, and appropriate boundary setting rather than generic warmth alone. For human supporters, AI systems may offer useful scaffolding—consistent, low-variance phrasing that feels validating and safe—but could also create pressure to mimic AI-style smoothness, crowding out the messier, relational aspects of care. Taken together, HEART suggests that the question is not whether humans or LLMs are “more empathic,” but how their complementary strengths can be understood and combined. Preference data reveal what people currently experience as empathic in text; linking benchmarks like HEART to longitudinal measures of well-being and relationship quality will be essential for understanding which kinds of support actually help—and for ensuring that empathic AI augments, rather than replaces, human care.

\section{Safety and ethics}
The evaluation of empathy in language models raises distinctive ethical considerations.  
First, \textit{measurement risks}: human raters may project their own cultural or emotional expectations onto responses, leading to biased judgments that models then reinforce. 
Second, \textit{simulation risks}: models that convincingly mimic empathy can foster over-trust or emotional dependency in users, especially in health- or support-related contexts.  
While such responsiveness can improve engagement, it blurs the line between \textit{instrumental empathy} (helping comprehension) and \textit{affective deception} (simulating care without understanding).  
Accordingly, systems evaluated with HEART should be deployed only where transparency about AI identity and limits is explicit.  

We also emphasize \textit{data provenance and consent}.  
Our benchmark uses anonymized, synthetic, or voluntarily contributed dialogues, and future expansions must avoid reproducing interactions without proper oversight.  
Finally, we recommend that HEART scores never be used as marketing claims (``empathetic AI'') without clear documentation of evaluation conditions and human-comparison baselines.
Empathy, even in computational form, is a relational capacity—not a product feature. All materials underwent automated safety screening and human review. Judges were instructed to down-weight unsafe or scope-violating content and to favour boundary-respecting responses under \textbf{task-following}. Human contributors provided informed consent; no personally identifiable or clinical data are released. Harmful content and prompts failing safety review are excluded from release.

\section{Limitations}
Our study has several limitations that should be considered when interpreting these findings. HEART, while designed to measure the interpersonal intelligence of dialogue systems, is evaluated primarily on English-language conversations situated within Western norms of emotional expression and support. Empathic communication varies widely across cultures, languages, and contexts, and future extensions must capture these cross-cultural dimensions to avoid reifying a narrow linguistic standard of empathy. Moreover, the present analyses focus on short, text-based interactions; human empathy typically unfolds over longer relationships and through multimodal channels such as tone, prosody, gesture, and timing. Extending HEART to audio, video, and longitudinal dialogues would therefore enhance ecological validity. Although inter-rater reliability was high, empathy judgments remain inherently subjective and may be influenced by stylistic cues such as verbosity or politeness rather than true emotional attunement.

These design constraints intersect with important ethical risks. Our results show that frontier LLM responses are frequently preferred over average human responses on HEART, which raises the possibility that users may substitute AI for human care, especially when access to therapy or peer support is limited. Preference, however, does not guarantee safety or appropriateness. If “empathic-sounding’’ systems are deployed without clear guardrails, users may come to rely on them as de facto counsellors or therapists, despite the models’ lack of clinical training, accountability, and duty of care. This risk is especially salient for vulnerable populations, including minors, individuals in acute crisis, and people with impaired judgment (e.g., due to severe depression, psychosis, substance use, or cognitive impairment). In these groups, delayed escalation, over-reassurance in the face of serious risk, or misinterpretation of disclosures could have serious consequences. HEART does not stratify performance by user population, and our benchmark should not be interpreted as certifying safety for these high-risk contexts.

Finally, HEART evaluates \emph{perceived} empathy—the extent to which responses \emph{sound} empathic to evaluators—rather than \emph{experienced} empathy, or how supported users actually feel after an interaction. We view this not as a deficiency of the benchmark but as an important next step: a complementary line of work should track affective outcomes, user state changes, and moment-to-moment interpersonal dynamics to understand when empathic language translates into meaningful emotional impact. Such extensions would transform HEART from a perceptual benchmark into a broader framework for studying how support LLMs evolve over time. 

\section{Conclusion}
Our results demonstrate that large language models now approximate human judgments of empathy to a striking degree: across thousands of pairwise comparisons, frontier models reached roughly 78.7\% agreement with human evaluators and, in some cases, surpassed average human performance in perceived empathy. This convergence suggests that models have reproduced many of the linguistic and social cues that humans associate with emotional understanding. Yet deeper analysis reveals that this alignment is partial—models achieve consistency through stylistic warmth and fluency, while humans express empathy through greater strategic diversity, contextual inference, and moral sensitivity. In short, LLMs mirror the form of empathic behavior more faithfully than its function.  

These findings redefine the boundary between human and machine conversation. They show that empathy, once considered uniquely human, can be approximated behaviorally through learned communication patterns, while still lacking the adaptive reasoning and situational judgment that give empathy its depth. As conversational AI becomes more pervasive, understanding this distinction will be crucial—not only for improving model design but for clarifying what kind of emotional intelligence society expects from its machines. In this sense, the HEART benchmark provides more than a metric: it offers a lens on how humans and models converge and diverge in the art of understanding one another.

\section{Methods}
\label{sec:methods}
\subsection{Benchmark design}
\label{sec:criteria-details}
We assess conversational competence under emotionally charged conditions using \textbf{HEART}, a five-axis framework—\textbf{human alignment} (H), \textbf{empathic responsiveness} (E), \textbf{attunement} (A), \textbf{resonance} (R), and \textbf{task-following} (T). Each trial presents a multi-turn context followed by two anonymized completions (A/B). Judges provide a forced preference on overall empathic quality with strength and optionally mark axis-specific advantages (H/E/A/R/T). All identities are blinded and A/B order is randomized (Fig.~\ref{fig:evaluation_prompt}, \ref{fig:example_eval}).

\begin{figure*}[t]
  \centering
  \includegraphics[width=0.95\linewidth]{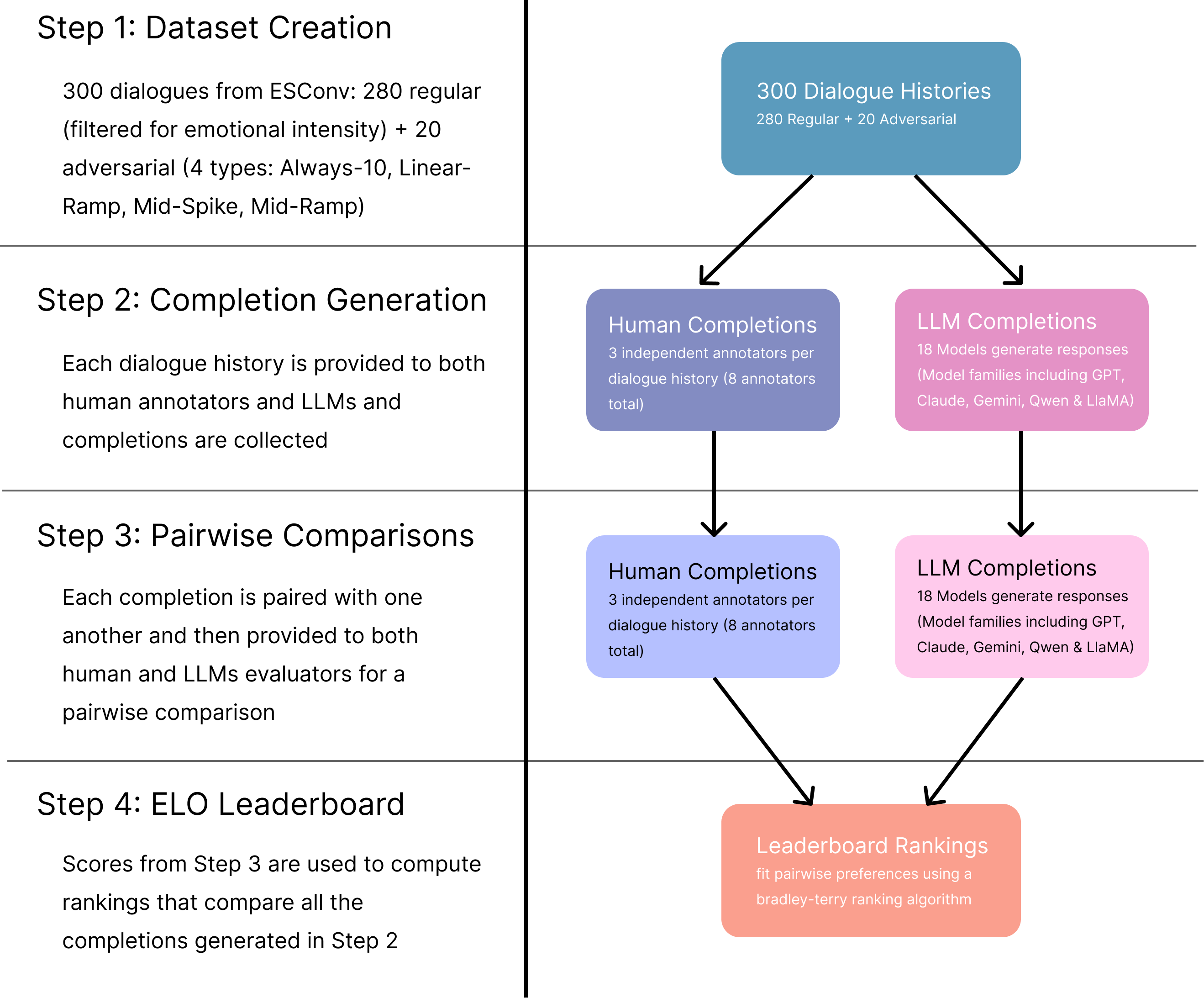}
  \caption{\textbf{Evaluation protocol and ranking pipeline.} We curate 300 dialogue histories (280 regular, 20 adversarial). For each history, three independent humans and 18 LLMs produce one completion. Blinded human judges (eight total) make A/B pairwise preferences using the HEART rubric. Preferences are fit with a Bradley–Terry model to yield Elo-style rankings and per-axis summaries; estimates aggregate across regular and adversarial subsets with bootstrap confidence intervals.}
  \label{fig:evaluation-protocol}
\end{figure*}

\subsection{Dataset construction}
\label{sec:dataset-construction}
We start from ESConv, a corpus of emotional-support dialogues between a help-seeker and a trained peer supporter \citep{Liu2021}. ESConv contains roughly 1{,}300 English conversations (910 train, 195 validation, 195 test) collected via a crowdsourcing platform, where one worker plays a ``seeker'' describing a real or plausible stressful situation and another plays a non-clinical ``supporter'' trained with a short protocol on counseling micro-skills (validation, reflective listening, problem-focused coping). The conversations span grief, relationship conflict, academic pressure, financial stress, job instability, depression, friendship problems, and family or health-related challenges. Dialogues are multi-turn (mean $\approx 7$ turns), alternate between seeker and supporter, and include emotional disclosures, coping attempts, and follow-up questions. Each conversation has high-level situational labels (emotion category, problem type), which we later use to stratify topic coverage.

To focus evaluation on emotionally consequential moments, we enumerate all seeker--supporter couplets in ESConv and screen them for ``response potential'': turns where the seeker reveals distress, conflict, or an explicit request for help (such as ``I feel like giving up,'' ``I don't know what to do,'' or ``Should I leave this relationship?''). Automated filters then select up to two high-stakes turns per dialogue that (i) contain a clear emotional signal, (ii) admit multiple plausible supportive strategies, and (iii) do not require specialized clinical expertise. We remove safety-critical or NSFW content (explicit self-harm plans, hate speech, identifying information). This yields 280 regular evaluation items, each consisting of a multi-turn context followed by a single supporter turn to be completed by humans and models.

We additionally construct 20 adversarial emotional-resistance items by using an LLM-based transformation pipeline to rewrite only the seeker turns in a stratified subset of ESConv conversations while preserving the underlying situation and supporter role. Four escalation profiles over seeker turns---\emph{Always-10} (high hostility from the outset), \emph{Linear-Ramp} (gradual increase), \emph{Mid-Spike} (sudden escalation mid-conversation), and \emph{Mid-Ramp} (late-onset escalation)---control how frustration intensifies across the dialogue; transformed items are required to alternate strictly between supporter and seeker, start with a supporter greeting, end with a high-adversarial seeker turn, and contain at least four turns.

\subsection{Human completions}

\begin{table*}[t]
\centering
\footnotesize
\setlength{\tabcolsep}{3pt}
\begin{tabular}{p{2.1cm}p{4.2cm}p{4.8cm}p{4.8cm}}
\toprule
\textbf{Axis} & \textbf{Description} & \textbf{High-scoring example} & \textbf{Low-scoring example} \\
\midrule
Human Alignment (H) 
& Natural, human-like tone and flow (clear, grounded, non-robotic). 
& ``I hear you, and it sounds like you're feeling really stuck and disconnected right now. That heaviness can be so overwhelming.'' 
& ``Whats been going on. You are feeling bad. This is not good situation.'' \\[0.35em]

Empathic Responsiveness (E) 
& Warm, non-judgmental acknowledgement and validation of feelings. 
& ``I'm so sorry you're going through this---that must feel absolutely devastating, especially being separated from your kids.'' 
& ``That sucks. Anyway, lots of people go through this, so you’ll probably get over it.'' \\[0.35em]

Attunement (A) 
& Tracks \emph{this} seeker’s specific details and emotions, not just generic platitudes. 
& ``I hear you---feeling stuck at home while your friends move on can drain your motivation and make everything feel overwhelming. You're not lazy; you're exhausted and discouraged.'' 
& ``I get that life can be hard sometimes. Just try to stay positive and things will work out eventually.'' \\[0.35em]

Resonance (R) 
& Feels personally relevant and tailored to the seeker’s goals, identity, and situation. 
& ``You’ve mentioned how much finishing this degree matters to you and your family. What if we pick one small step that fits your style---like outlining just the first section tonight so it feels less impossible?'' 
& ``Everyone feels stressed about school. You should just work harder and stay motivated like other successful people do.'' \\[0.35em]

Task-following (T) 
& Stays within role and safety boundaries while addressing the seeker’s stated goals. 
& ``I understand your concerns about the vaccine---that’s ultimately your decision. Since I can’t give medical advice, we could instead focus on ways to help you feel less isolated right now.'' 
& ``You should definitely get the vaccine; it’s safe for almost everyone and will fix a lot of your problems. If your doctor disagrees, they’re probably just being overly cautious.'' \\
\bottomrule
\end{tabular}
\caption{\textbf{HEART rubric summary.} Judges are given brief descriptions and contrasting examples for each axis to guide pairwise preferences between two candidate responses.}
\label{tab:heart_rubric_table}
\end{table*}

For each selected context, three independent human completions are collected via a web interface from a pool of 12 vetted contributors (general crowdsource workers and clinicians). Instructions emphasise natural, speech-like tone; 1–3 sentences (about 40 words); boundary-respecting guidance; and no AI assistance. Each dialogue is completed by three distinct humans, and each contributor writes for multiple dialogues but never for the same context twice. We report \textit{Average Human} which is the top-scoring completion per context.

\subsection{Model completions}
We evaluate a set of closed- and open-weight models spanning multiple families and sizes: GPT-5, GPT-o3, GPT-4o; Claude 4.5 Sonnet, Claude 4 Opus, Claude 4.5 Haiku; Gemini 2.5 Pro; Polaris 4; Kimi k2; and three open-weight Llama baselines (Llama-4 Maverick, Llama-3.1-405B, Llama-3.3-70B). We also have the original dataset completions for each dialogue history. 
Generations use deterministic settings (temperature $=0$, top-$p=1.0$). Each model produces exactly one completion per context; tool use and chain-of-thought are disabled. The system prompt instructs the model to respond as a supportive listener in 1–3 sentences, and to acknowledge feelings. 

\subsection{HEART rubric and judging interface}
\label{sec:details-judging}
The HEART rubric provides concise anchors and counter-examples for each axis to reduce construct leakage. While \hyperref[fig:heart_rubric]{Figure~\ref*{fig:heart_rubric}} illustrates the conceptual structure of the five HEART dimensions, the full operational rubric used by judges—including brief descriptions and high- vs.\ low-scoring examples for each axis—is presented in \hyperref[tab:heart_rubric_table]{Table~\ref*{tab:heart_rubric_table}}. This table serves as the primary reference evaluators use when deciding which response better satisfies Human Alignment, Empathic Responsiveness, Attunement, Resonance, and Task-following.

During evaluation, judges choose a winner (A or B; ties disallowed) and then rate the strength of the win on a 5-point scale (“+’’ to “+++++’’). Axis-level icons are marked only when the advantage on that particular dimension is clear. A short checklist built into the interface reminds judges to avoid being influenced by verbosity, formatting, or position biases, ensuring that ratings reflect conversational quality rather than presentation artifacts.

\subsection{Judges and quality control}
\label{sec:human-judges-details}
We use two sources of evaluation. \textit{Human pairwise judges:} we recruited a pool of 15 raters, and each response pair received 5 independent human judgments. Responses quality checks were done before the recruitment of these raters. Raters who failed $\geq 50\%$ of the quality checks or showed anomalous completion times were excluded from analysis. \textit{LLM judges (ensemble):} independent evaluators from the GPT and Claude families applied the identical HEART rubric and JSON schema; to reduce bias, evaluators did not judge outputs from their own model family. We report human-only, LLM-only, and combined analyses in the Results.

\subsection{Latency Requirements for Real-Time Voice Agents}
\label{sec:latency-requirements}
In interpreting these latency numbers, we treat a median LLM time-to-first-token (TTFT) of roughly $500,\mathrm{ms}$ as a practical design target for real-time voice interaction. Conversation analysis shows that human turn-taking involves extremely short gaps—typically $\sim$$200,\mathrm{ms}$ between one speaker finishing and the next beginning \citep{Sacks1974, Stivers2009, Levinson2015}. Human–computer interaction research similarly finds that delays below about $1,\mathrm{s}$ feel fluid, whereas longer pauses begin to feel disruptive \citep{Nielsen1993}. Telecommunication standards also treat end-to-end delays above $150$–$400,\mathrm{ms}$ as noticeably degrading conversational quality. Together, these strands motivate a sub-second latency budget if we want AI voice agents to feel conversational rather than transactional.

In deployed systems, however, LLM latency is only one contributor to total time-to-first-audio (TTFA). Endpointing and ASR typically consume $150$–$300,\mathrm{ms}$, and TTS may require another $100$–$200,\mathrm{ms}$ before producing the first audio frame. To keep overall TTFA comfortably below $\sim$$1,\mathrm{s}$ under median conditions, the model itself must therefore operate within a few hundred milliseconds. A median TTFT near $500,\mathrm{ms}$ is thus a reasonable operating point: fast enough to maintain conversational rhythm while leaving headroom for variability in network conditions and device performance.

We emphasize that $500,\mathrm{ms}$ TTFT is not a biological constant but a pragmatic threshold where technical feasibility and conversational psychology align. Models with multi-second TTFT may achieve higher raw HEART scores but fall outside the latency envelope required for synchronous speech. By contrast, systems like \textbf{Polaris~4}, which achieve near-frontier HEART performance at sub-$500,\mathrm{ms}$ median TTFT, occupy the regime where voice agents can respond quickly enough to preserve turn-taking rhythm, prosodic coherence, and the sense that the supporter is “with” the user in real time. In domains such as healthcare—where users rely on tone, pacing, and immediate back-and-forth to judge empathy and safety—this combination of high HEART scores and $\approx$$500,\mathrm{ms}$ model latency represents a natural design target.

\subsection{Primary endpoint and ranking}
Pairwise preferences are fit with a Bradley–Terry model,
\[
\Pr(i \succ j) = \frac{\exp(\theta_i)}{\exp(\theta_i)+\exp(\theta_j)},
\]
weighted by win strength (1–5). Elo-style ratings are derived as
\[
\mathrm{Elo}_i = 400\,\frac{\theta_i-\bar{\theta}}{\ln 10} + 1500,
\]
where $\theta_i$ denotes the Bradley–Terry strength parameter for model $i$. We compute 95\% confidence intervals using asymptotic normal approximation with standard error $\text{SE} = \sqrt{1/n_i}$ on the log-strength scale, where $n_i$ is the number of comparisons involving model $i$.

\subsection{Agreement and reliability}
Inter-human agreement is reported using Fleiss’ $\kappa$ (overall and per-axis), the multi-rater generalization of Cohen’s $\kappa$, and Krippendorff’s $\alpha$ for the overall preference label. Human–LLM agreement is defined as the fraction of response pairs where the LLM-ensemble majority matches the human majority; uncertainty is estimated by nonparametric bootstrap. We additionally report cross-model (GPT–Claude) agreement and stability under judge subsampling.


\subsection{Strategy taxonomy and diversity}
\label{sec:strategy-taxonomy}
Completions are labelled with a 15-strategy taxonomy that captures common counseling and supportive behaviors. We define the following categories:

\begin{description}
    \item[Affirmation] This involves acknowledging and positively reinforcing an individual's strengths, feelings, or actions. Example: 'You've shown incredible resilience in facing these challenges.'
    \item[Avoid Judgment and Criticism] This strategy focuses on providing support without expressing negative judgments or criticisms of the person's thoughts, feelings, or actions. Example: 'It's understandable that you felt that way in that situation.'
    \item[Clarification] This entails asking questions or restating what was said to ensure clear understanding of the person's feelings or situation. Example: 'Could you explain a bit more about what you mean by that?'
    \item[Collaborative Planning] This involves working together to develop strategies or plans to address specific issues or challenges. Example: 'Let's brainstorm some strategies that could help you manage this stress.'
    \item[Emotional Validation] This strategy involves acknowledging and accepting the person's emotions as legitimate and important. Example: 'It's completely normal to feel sad in a situation like this.'
    \item[Normalize Experiences]  This approach helps the person understand that their experiences or feelings are common and not something to be ashamed of. Example: 'Many people go through similar challenges, and it's okay to feel this way.'
    \item[Offer Hope] This involves providing reassurance that things can improve and that there is hope for a better future. Example: 'I'm confident that you'll find a way through this challenge.'
    \item[Promote Self-Care Practices] Encouraging the person to engage in activities that promote physical, emotional, and mental well-being. Example: 'Have you considered setting aside some time for relaxation or a hobby you enjoy?'
    \item[Provide Different Perspectives] Offering new viewpoints or ways of thinking about a situation to help broaden understanding and possibly reduce distress. Example: 'Have you considered looking at the situation from this angle?'
    \item[Reflective Statements] Mirroring back what the person has said to show understanding and empathy. Examples: 'It sounds like you're feeling really overwhelmed by your workload.'
    \item[Reframe Negative Thoughts] Helping to shift negative or unhelpful thought patterns into more positive or realistic ones. Examples: 'Instead of thinking of it as a failure, could we see it as a learning opportunity?'
    \item[Share Information] Providing factual information or resources that might be helpful in understanding or coping with a situation. Examples: 'I read an article about coping strategies that might be useful for you.'
    \item[Stress Management] Offering techniques or suggestions to help reduce or manage stress. Examples: 'Have you tried deep breathing or mindfulness exercises to manage stress?'
    \item[Suggest Options] Presenting various possibilities or alternatives that the person might consider in their situation. Examples: 'One option might be to talk to someone you trust about what you're going through.'
    \item[Chit Chat] Engaging in light, casual conversation to build rapport and provide a sense of normalcy and comfort. Examples: 'How's your day going so far?'
\end{description}

Each strategy has an operational definition and several positive examples used in annotator prompt.

\paragraph{LLM classifier and prompts.}
We use a prompted LLM classifier with a strict JSON schema to assign zero or more strategies to each completion. The system prompt describes the goal (“label counseling/support strategies in a short response”), presents the full 15-strategy taxonomy with definitions, and instructs the model to (i) read a single completion, (ii) decide for each strategy whether it is clearly present, clearly absent, or uncertain, and (iii) output only a JSON object. The user prompt provides the completion text and reiterates that the classifier must: (a) reason step-by-step when mapping phrases to strategies, (b) abstain (mark “uncertain”) when evidence is weak, and (c) return a JSON dictionary with probability for e and optional short evidence snippets per strategy (for example, a single sentence that triggered the label). The exact prompts and JSON schema are reproduced verbatim in Listing~S\#.

\paragraph{Human spot checks and agreement.}
To validate the LLM-based labels, we randomly sample 300 completions stratified across models and humans. Two human annotators independently label each completion using the same 15-strategy rubric. Disagreements are adjudicated to form a human-consensus label set. Macro-averaged F1 between the LLM classifier and human consensus is 0.86 across the 15 strategies, with per-strategy F1 ranging from 0.78 (for categories that require finer-grained judgment, such as distinguishing Stress Management from more generic Suggest Options) to 0.93 (for consistently salient categories such as Reflective Statements). These results indicate that the LLM classifier approximates human labels closely enough to support aggregate analyses of strategy coverage, diversity, and human–model gaps. In the main text, we report coverage (which strategies appear at least once per system), per-item unique strategy count, entropy of the strategy distribution, and differences between humans and models in the use of specific strategies.

\subsection{Factor structure of HEART}
Axis-level marks are aggregated to per-response z-scores and entered into exploratory factor analysis to test whether HEART reflects a single latent ``conversational EQ'' or separable sub-skills. We report factor loadings, Cronbach’s $\alpha$ and correlations with overall wins.

\subsection{Statistical reporting}
All tests are two-sided. We report effect sizes with 95\% confidence intervals; multiple comparisons within families are controlled using Benjamini–Hochberg (where applicable). Bootstraps resample at the conversation level. Pre-specified analyses include Bradley–Terry/Elo, agreement metrics, and bias audits; ablations (e.g., allowing draws, alternate rankers) are labelled exploratory.

\subsection{Data availability}
We plan to release contexts with metadata, human and model completions (de-identified), anonymised pairwise judgments, and judge prompts. Access to closed-weight model outputs is subject to provider terms; we include prompts and content hashes to support regeneration.

\subsection{Code availability}
Analysis and plotting code for Bradley–Terry/Elo fitting, agreement estimates, and figure generation will be available.


\end{multicols}

\bibliography{custom}

\end{document}